\documentclass[11pt,a4paper]{article}
\usepackage[hyperref]{naaclhlt2019}
\usepackage{times}
\usepackage{latexsym}

\usepackage{amsfonts}
\usepackage{url}
\usepackage{xcolor}
\usepackage[ruled,linesnumbered]{algorithm2e}
\usepackage{graphicx}
\usepackage{bbm}

\usepackage{amsmath}
\DeclareMathOperator*{\argmax}{arg\,max}

\aclfinalcopy 

\title{BERT Post-Training for Review Reading Comprehension\\and Aspect-based Sentiment Analysis}

\author{Hu Xu\textsuperscript{\text{1}}, Bing Liu\textsuperscript{\text{1}}, Lei Shu\textsuperscript{\text{1}}\and Philip S. Yu\textsuperscript{\text{1,2}}\\
    \textsuperscript{1}{Department of Computer Science, University of Illinois at Chicago, Chicago, IL, USA}\\
    \textsuperscript{2}{Institute for Data Science, Tsinghua University, Beijing, China}\\
    {\tt \{hxu48, liub, lshu3, psyu\}@uic.edu}
}

\date{}

\begin{document}
\maketitle
\begin{abstract}
Question-answering plays an important role in e-commerce as it allows potential customers to actively seek crucial information about products or services to help their purchase decision making. 
Inspired by the recent success of machine reading comprehension (MRC) on formal documents, this paper explores the potential of turning customer reviews into a large source of knowledge that can be exploited to answer user questions.~We call this problem \textit{\underline{R}eview \underline{R}eading \underline{C}omprehension} (RRC).~To the best of our knowledge, no existing work has been done on RRC. In this work, we first build an RRC dataset called ReviewRC based on a popular benchmark for aspect-based sentiment analysis.~Since ReviewRC has limited training examples for RRC (and also for aspect-based sentiment analysis), we then explore a novel post-training approach on the popular language model BERT to enhance the performance of fine-tuning of BERT for RRC.
To show the generality of the approach, the proposed post-training is also applied to some other review-based tasks such as aspect extraction and aspect sentiment classification in aspect-based sentiment analysis. 
Experimental results demonstrate that the proposed post-training is highly effective\footnote{The datasets and code are available at \url{https://www.cs.uic.edu/~hxu/}. 
}.
\end{abstract}

\section{Introduction}

For online commerce, question-answering (QA) serves either as a standalone application of customer service or as a crucial component of a dialogue system that answers user questions.
Many intelligent personal assistants (such as Amazon Alexa and Google Assistant) support online shopping by allowing the user to speak directly to the assistants. 
One major hindrance for this mode of shopping is that such systems have limited capability to answer user questions about products (or services), which are vital for customer decision making.
As such, an intelligent agent that can automatically answer customers' questions is very important for the success of online businesses.

Given the ever-changing environment of products and services, it is very hard, if not impossible, to pre-compile an up-to-date and reliable knowledge base to cover a wide assortment of questions that customers may ask, such as in factoid-based KB-QA \cite{xu2016question,fader2014open,kwok2001scaling,yin2015neural}.
As a compromise, many online businesses leverage community question-answering (CQA) \cite{mcauley2016addressing} to crowdsource answers from existing customers. However, the problem with this approach is that many questions are not answered, and if they are answered, the answers are delayed, which is not suitable for interactive QA.
In this paper, we explore the potential of using product reviews as a large source of user experiences that can be exploited to obtain answers to user questions. Although there are existing studies that have used information retrieval (IR) techniques \cite{mcauley2016addressing,yu2018aware} to find a whole review as the response to a user question, giving the whole review to the user is undesirable as it is quite time-consuming for the user to read it.

Inspired by the success of \underline{M}achine \underline{R}eading \underline{C}omphrenesions (MRC) \cite{rajpurkar2016squad,rajpurkar2018know}, we propose a novel task called \underline{R}eview \underline{R}eading \underline{C}omprehension (RRC) as following.

\textbf{Problem Definition}: Given a question $q=(q_1, \dots, q_m)$ from a customer (or user) about a product and a review $d=(d_1, \dots, d_n)$ for that product containing the information to answer $q$, find a sequence of tokens (a text span) $a=(d_s, \dots, d_e)$ in $d$ that answers $q$ correctly, where $1 \le s \le n$, $1\le e \le n$, and $s\le e$.

\label{sec:intro}
\begin{table}
    \centering
    \scalebox{0.87}{
        \begin{tabular}{|l|}
            \hline
            {\bf Questions}\\
            \hline
            Q1: Does it have an internal hard drive ?\\
            Q2: How large is the internal hard drive ?\\
            Q3: is the capacity of the internal hard drive OK ?\\
            \hline
            {\bf Review}\\
            Excellent value and a must buy for someone \\
            looking for a Macbook . You ca n't get any \\
            better than this price and it \textbf{come with}\textsubscript{A1} an\\
            internal disk drive . All the newer MacBooks\\
            do not . Plus you get \textbf{500GB}\textsubscript{A2} which is also a\\
            \textbf{great}\textsubscript{A3} feature . Also , the resale value on \\
            this will keep . I highly recommend you get one \\
            before they are gone .\\
            \hline
        \end{tabular}
    }
	\caption{An example of review reading comprehension: we show 3 questions and their corresponding answer spans from a review.}
    \label{tbl:example}
\end{table}

A sample \emph{laptop} review is shown in Table \ref{tbl:example}. 
We can see that customers may not only ask factoid questions such as the specs about some aspects of the laptop as in the first and second questions but also subjective or opinion questions about some aspects (capacity of the hard drive), as in the third question.
RRC poses some \textit{domain challenges} compared to the traditional MRC on Wikipedia, such as the need for rich product knowledge, informal text, and fine-grained opinions (there is almost no subjective content in Wikipedia articles). Research also shows that yes/no questions are very frequent for products with complicated specifications \cite{mcauley2016addressing,Xu2018pro}.

To the best of our knowledge, no existing work has been done in RRC. This work first builds an RRC dataset called ReviewRC, using reviews from SemEval 2016 Task 5\footnote{\url{http://alt.qcri.org/semeval2016/task5/}. We choose these review datasets to align RRC with existing research on sentiment analysis.}, which is a popular dataset for aspect-based sentiment analysis (ABSA) \cite{hu2004mining} in the domains of \emph{laptop} and \emph{restaurant}.
We detail ReviewRC in Sec. \ref{sec:exp}.
Given the wide spectrum of domains (types of products or services) in online businesses and the prohibitive cost of annotation, ReviewRC can only be considered to have a limited number of annotated examples for supervised training, which still leaves the domain challenges partially unresolved.

This work adopts BERT 
\cite{devlin2018bert} as the base model as it achieves the state-of-the-art performance on MRC \cite{rajpurkar2016squad,rajpurkar2018know}.
Although BERT aims to learn contextualized representations across a wide range of NLP tasks (to be task-agnostic), leveraging BERT alone still leaves the domain challenges unresolved (as BERT is trained on Wikipedia articles and has almost no understanding of opinion text), and it also introduces another challenge of task-awareness (the RRC task), called the \textit{task challenge}.~This challenge arises when the task-agnostic BERT meets the limited number of fine-tuning examples in ReviewRC (see Sec.~\ref{sec:exp}) for RRC, which is insufficient to fine-tune BERT to ensure full task-awareness of the system\footnote{The end tasks from the original BERT paper typically use tens of thousands of examples to ensure that the system is task-aware.}. 
To address all the above challenges,
we propose a novel joint post-training technique that takes BERT's pre-trained weights as the initialization\footnote{Due to limited computation resources, it is impractical for us to pre-train BERT directly on reviews from scratch \cite{devlin2018bert}.} for basic language understanding and adapt BERT with both domain knowledge and task (MRC) knowledge before fine-tuning using the domain end task annotated data for the domain RRC.
This technique leverages knowledge from two sources: unsupervised domain reviews and supervised (yet out-of-domain) MRC data \footnote{To simplify the writing, we refer MRC as a general-purpose RC task on formal text (non-review) and RRC as an end-task specifically focused on reviews.}, where the former enhances domain-awareness and the latter strengthens MRC task-awareness.
As a general-purpose approach, we show that the proposed method can also benefit ABSA tasks such as aspect extraction (AE) and aspect sentiment classification (ASC).

The main contributions of this paper are as follows.
(1) It proposes the new problem of \textit{review reading comprehension} (RRC).
(2) To solve this new problem, an annotated dataset for RRC is created.
(3) It proposes a general-purpose post-training approach to improve RRC, AE, and ASC.
Experimental results demonstrate that the proposed approach is effective. 

\section{Related Works}
\label{sec:rel}

Many datasets have been created for MRC from formally written and objective texts, e.g., Wikipedia (WikiReading \cite{hewlett2016wikireading}, SQuAD \cite{rajpurkar2016squad,rajpurkar2018know}, WikiHop \cite{welbl2018constructing}, DRCD \cite{shao2018drcd}, QuAC \cite{choi2018quac}, HotpotQA \cite{yang2018hotpotqa}) news and other articles (CNN/Daily Mail \cite{hermann2015teaching}, NewsQA \cite{trischler2016newsqa}, RACE \cite{lai2017race}), fictional stories (MCTest \cite{richardson2013mctest}, CBT \cite{hill2015goldilocks}, NarrativeQA \cite{kovcisky2018narrativeqa}), and general Web documents (MS MARCO \cite{nguyen2016ms}, TriviaQA \cite{joshi2017triviaqa}, SearchQA \cite{dunn2017searchqa} ). 
Also, CoQA \cite{reddy2018coqa} is built from multiple sources, such as Wikipedia, Reddit, News, Mid/High School Exams, Literature, etc.
To the best of our knowledge, MRC has not been used on reviews, which are primarily subjective. As such, we created a review-based MRC dataset called ReviewRC.
Answers from ReviewRC are extractive (similar to SQuAD \cite{rajpurkar2016squad,rajpurkar2018know}) rather than abstractive (or generative) (such as in MS MARCO \cite{nguyen2016ms} and CoQA \cite{reddy2018coqa}).
This is crucial because online businesses are typically cost-sensitive and extractive answers written by humans can avoid generating incorrect answers beyond the contents in reviews by an AI agent.

Community QA (CQA) is widely adopted by online businesses \cite{mcauley2016addressing} to help users.
However, since it solely relies on humans to give answers, it often takes a long time to get a question answered or even not answered at all as we discussed in the introduction.
Although there exist researches that align reviews to questions as an information retrieval task \cite{mcauley2016addressing,yu2018aware}, giving a whole review to the user to read is time-consuming and not suitable for customer service settings that require interactive responses.

Knowledge bases (KBs) (such as Freebase \cite{dong2015question,xu2016question,yao2014information} or DBpedia \cite{lopez2010scaling,unger2012template}) have been used for question answering \cite{yu2018aware}.
However, the ever-changing nature of online businesses, where new products and services appear constantly, makes it prohibitive to build a high-quality KB to cover all new products and services.

Reviews also serve as a rich resource for sentiment analysis \cite{pang2002thumbs,hu2004mining,liu2012sentiment,liu2015sentiment}.
Although document-level (review) sentiment classification may be considered as a solved problem (given ratings are largely available), aspect-based sentiment analysis (ABSA) is still an open challenge, where alleviating the cost of human annotation is also a major issue.
ABSA aims to turn unstructured reviews into structured fine-grained aspects (such as the ``battery'' of a laptop) and their associated opinions (e.g., ``good battery'' is \emph{positive} about the aspect battery).
Two important tasks in ABSA are aspect extraction (AE) and aspect sentiment classification (ASC) \cite{hu2004mining}, where the former aims to extract aspects (e.g., ``battery'') and the latter targets to identify the polarity for a given aspect (e.g., \emph{positive} for \emph{battery}).
Recently, supervised deep learning models dominate both tasks \cite{wang2016recursive,wang2017coupled,xu_acl2018,tang2016aspect,he2018exploiting} and many of these models use handcrafted features, lexicons, and complicated neural network architectures to remedy the insufficient training examples from both tasks.
Although these approaches may achieve better performances by manually injecting human knowledge into the model, human baby-sat models may not be intelligent enough\footnote{\url{http://www.incompleteideas.net/IncIdeas/BitterLesson.html}} and automated representation learning from review corpora is always preferred \cite{xu_acl2018,he2018exploiting}.
We push forward this trend with the recent advance in pre-trained language models from deep learning~\cite{peters2018deep,howard2018universal,devlin2018bert,radford2018improving,radford2018lang}. 
Although it is practical to train domain word embeddings from scratch on large-scale review corpora \cite{xu_acl2018}, it is impractical to train language models from scratch with limited computational resources.
As such, we show that it is practical to adapt language models pre-trained from formal texts to domain reviews.

\section{BERT and Review-based Tasks}
In this section, we briefly review BERT and derive its fine-tuning formulation on three (3) review-based end tasks.

\subsection{BERT}
BERT is one of the key innovations in the recent progress of contextualized representation learning \cite{peters2018deep,howard2018universal,radford2018improving,devlin2018bert}.
The idea behind the progress is that even though the word embedding \cite{mikolov2013distributed,pennington2014glove} layer (in a typical neural network for NLP) is trained from large-scale corpora, training a wide variety of neural architectures that encode contextual representations only from the limited supervised data on end tasks is insufficient.
Unlike ELMo \cite{peters2018deep} and ULMFiT \cite{howard2018universal} that are intended to provide additional features for a particular architecture that bears human's understanding of the end task, BERT adopts a fine-tuning approach that requires almost no specific architecture for each end task. This is desired as an intelligent agent should minimize the use of prior human knowledge in the model design. Instead, it should learn such knowledge from data. BERT has two parameter intensive settings: 

\noindent
$\textbf{BERT}_\textbf{BASE}$: 12 layers, 768 hidden dimensions and 12 attention heads (in transformer) with the total number of parameters, 110M;

\noindent
$\textbf{BERT}_\textbf{LARGE}$: 24 layers, 1024 hidden dimensions and 16 attention heads (in transformer) with the total number of parameters, 340M.

\begin{figure}[t]
\centering    
\includegraphics[width=3.0in]{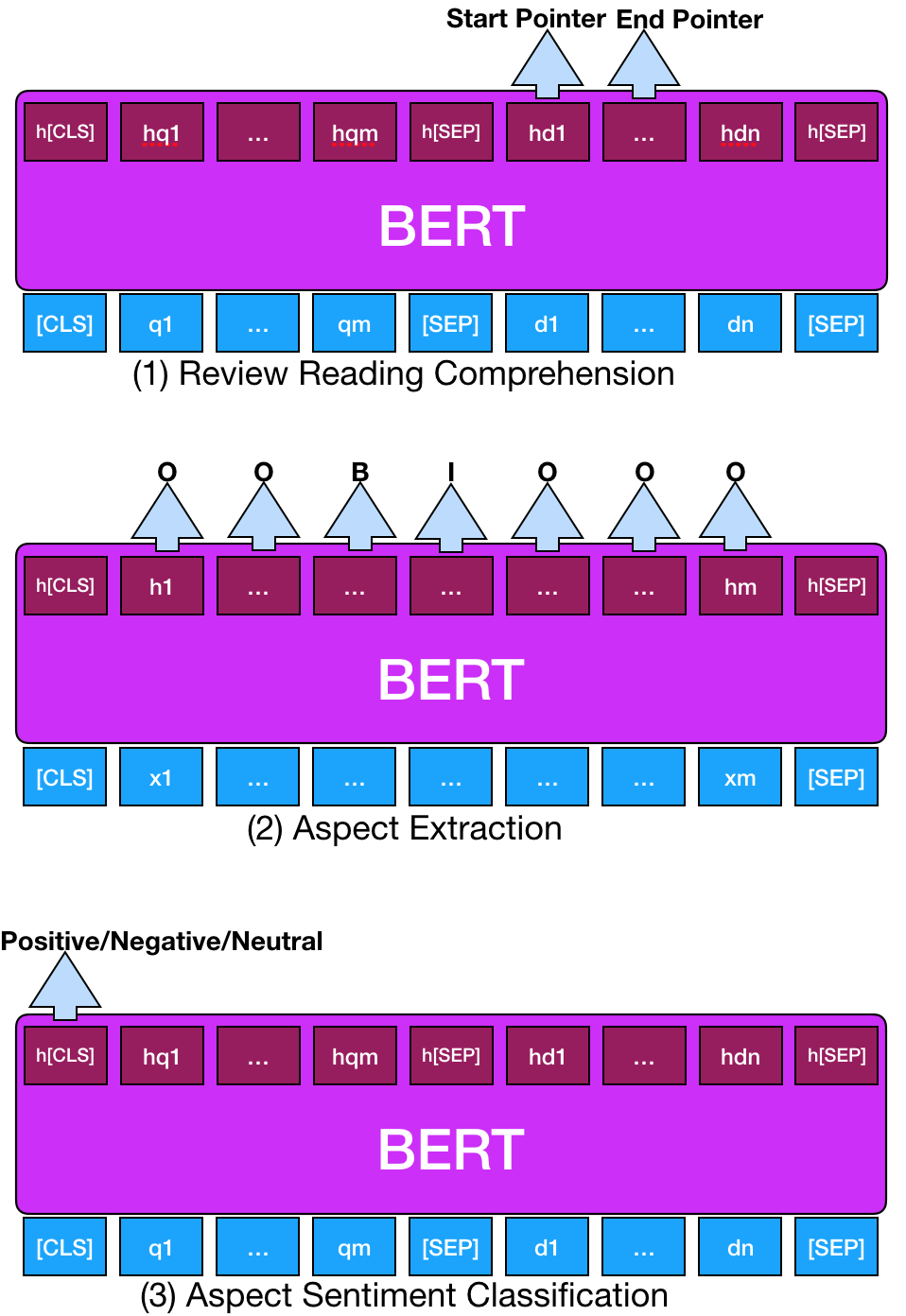}
    \caption{Overview of BERT settings for review reading comprehension (RRC), aspect extraction (AE) and aspect sentiment classification (ASC).}
\label{fig:overview}
\vspace{-3mm}
\end{figure}

We only extend BERT with one extra task-specific layer and fine-tune BERT on each end task.
We focus on three (3) review-based tasks: review reading comprehension (RRC), aspect extraction (AE) and aspect sentiment classification (ASC). The inputs/outputs settings are depicted in Figure \ref{fig:overview} and detailed in the following subsections.

\subsection{Review Reading Comprehension (RRC)}
\label{sec:rrc}
Following the success of SQuAD \cite{rajpurkar2016squad} and BERT's SQuAD implementation, we design review reading comprehension as follows.
Given a question $q=(q_1, \dots, q_m)$ asking for an answer from a review $d=(d_1, \dots, d_n)$, we formulate the input as a sequence $x=(\texttt{[CLS]}, q_1, \dots, q_m, \texttt{[SEP]}, d_1, \dots, d_n, \texttt{[SEP]})$, where \texttt{[CLS]} is a dummy token not used for RRC and \texttt{[SEP]} is intended to separate $q$ and $d$.
Let $\text{BERT}(\cdot)$ be the pre-trained (or post-trained as in the next section) BERT model. We first obtain the hidden representation as $h=\text{BERT}(x) \in \mathbb{R}^{r_h*|x|}$, where $|x|$ is the length of the input sequence and $r_h$ is the size of the hidden dimension. Then the hidden representation is passed to two separate dense layers followed by softmax functions: $l_1=\text{softmax}(W_1 \cdot h + b_1)$ and $l_2=\text{softmax}(W_2 \cdot h + b_2)$, where $W_1$, $W_2 \in \mathbb{R}^{r_h}$ and $b_1, b_2 \in \mathbb{R}$. The \text{softmax} is applied along the dimension of the sequence.
The output is a span across the positions in $d$ (after the \texttt{[SEP]} token of the input), indicated by two pointers (indexes) $s$ and $e$ computed from $l_1$ and $l_2$: $s=\argmax_{ \text{Idx}_{\texttt{[SEP]}} < s<|x|}(l_1)$ and $e=\argmax_{s\le e<|x|}(l_2)$, where $\text{Idx}_{\texttt{[SEP]}}$ is the position of token \texttt{[SEP]} (so the pointers will never point to tokens from the question).
As such, the final answer will always be a valid text span from the review as $a=(d_s, \dots, d_e)$.

Training the RRC model involves minimizing the loss that is designed as the averaged cross entropy on the two pointers: $$\mathcal{L}_{\text{RRC}}=-\frac{\sum \log l_1 \mathbb{I}(s)+ \sum \log l_2 \mathbb{I}(e)}{2},$$ where $\mathbb{I}(s)$ and $\mathbb{I}(e)$ are one-hot vectors representing the ground truths of pointers.

RRC may suffer from the prohibitive cost of annotating large-scale training data covering a wide range of domains. 
And BERT severely lacks two kinds of prior knowledge: (1) large-scale domain knowledge (e.g., about a specific product category), and (2) task-awareness knowledge (MRC/RRC in this case).
We detail the technique of jointly incorporating these two types of knowledge in Sec. \ref{sec:pt}.

\subsection{Aspect Extraction}
As a core task in ABSA, aspect extraction (AE) aims to find aspects that reviewers have expressed opinions on \cite{hu2004mining}. 
In supervised settings, it is typically modeled as a sequence labeling task, where each token from a sentence is labeled as one of $\{\textit{\underline{B}egin}, \textit{\underline{I}nside}, \textit{\underline{O}utside}\}$. A continuous chunk of tokens that are labeled as one \textit{B} and followed by zero or more \textit{I}s forms an aspect.
The input sentence with $m$ words is constructed as $x=(\texttt{[CLS]}, x_1, \dots, x_m, \texttt{[SEP]})$.
After $h=\text{BERT}(x)$, we apply a dense layer and a softmax for each position of the sequence: $l_3=\text{softmax}(W_3 \cdot h + b_3)$, where $W_3 \in \mathbb{R}^{3*r_h}$ and $b_3 \in \mathbb{R}^3$ (3 is the total number of labels (\textit{BIO})).  Softmax is applied along the dimension of labels for each position and $l_3 \in [0, 1]^{3*|x|}$. The labels are predicted as taking argmax function at each position of $l_3$ and the loss function is the averaged cross entropy across all positions of a sequence.

AE is a task that requires intensive domain knowledge (e.g., knowing that ``screen'' is a part of a laptop). Previous study \cite{xu_acl2018} has shown that incorporating domain word embeddings greatly improve the performance. 
Adapting BERT's general language models to domain reviews is crucial for AE,
as shown in Sec. \ref{sec:exp}.

\subsection{Aspect Sentiment Classification}
As a subsequent task of AE, aspect sentiment classification (ASC) aims to classify the sentiment polarity (positive, negative, or neutral) expressed on an aspect extracted from a review sentence.
There are two inputs to ASC: an aspect and a review sentence mentioning that aspect.
Consequently, ASC is close to RRC as the question is just about an aspect and the review is just a review sentence but ASC only needs to output a class of polarity instead of a textual span.

Let $x=(\texttt{[CLS]}, q_1, \dots, q_m, \texttt{[SEP]}, d_1, \dots, \\ d_n, \texttt{[SEP]})$, where $q_1, \dots, q_m$ now is an aspect (with $m$ tokens) and $d_1, \dots, d_n$ is a review sentence containing that aspect.
After $h=\text{BERT}(x)$, we leverage the representations of \texttt{[CLS]} $h_{\text{[CLS]}}$, which is the aspect-aware representation of the whole input.
The distribution of polarity is predicted as $l_4=\text{softmax}(W_4 \cdot h_{\text{[CLS]}} + b_4)$, where $W_4 \in \mathbb{R}^{3*r_h}$ and $b_4 \in \mathbb{R}^3$ (3 is the number of polarities). Softmax is applied along the dimension of labels on \texttt{[CLS]}: $l_4 \in [0, 1]^{3}$.
Training loss is the cross entropy on the polarities.

As a summary of these tasks, insufficient supervised training data significantly limits the performance gain across these 3 review-based tasks.~Although BERT's pre-trained weights strongly boost the performance of many other NLP tasks on formal texts, we observe in Sec. \ref{sec:exp} that BERT's weights only result in limited gain or worse performance compared with existing baselines.
In the next section, we introduce the post-training step to boost the performance of all these 3 tasks.

\section{Post-training}
\label{sec:pt}
As discussed in the introduction, fine-tuning BERT directly on the end task that has limited tuning data faces both domain challenges and task-awareness challenge.
To enhance the performance of RRC (and also AE and ASC), we may need to reduce the bias introduced by non-review knowledge (e.g., from Wikipedia corpora) and fuse domain knowledge (DK) (from unsupervised domain data) and task knowledge (from supervised MRC task but out-of-domain data).
Given MRC is a general task with answers of questions covering almost all document contents, a large-scale MRC supervised corpus may also benefit AE and ASC.
Eventually, we aim to have a general-purpose post-training strategy that can exploit the above two kinds of knowledge for end tasks.

To post-train on domain knowledge, we leverage the two novel pre-training objectives from BERT: masked language model (MLM) and next sentence\footnote{The BERT paper refers a sentence as a piece of text with one to many natural language sentences.} prediction (NSP). The former predicts randomly masked words and the latter detects whether two sides of the input are from the same document or not.~A training example is formulated as $(\texttt{[CLS]}, x_{1:j}, \texttt{[SEP]}, x_{j+1:n}, \texttt{[SEP]})$, where $x_{1:n}$ is a document (with randomly masked words) split into two sides $x_{1:j}$ and $x_{j+1:n}$ and \texttt{[SEP]} separates those two.

MLM is crucial for injecting review domain knowledge and for alleviating the bias of the knowledge from Wikipedia. 
For example, in the Wikipedia domain, BERT may learn to guess the \texttt{[MASK]} in ``The \texttt{[MASK]} is bright'' as ``sun''. But in a laptop domain, it could be ``screen''.
Further, if the \texttt{[MASK]}ed word is an opinion word in ``The touch screen is \texttt{[MASK]}'', this objective challenges BERT to learn the representations for fine-grained opinion words like ``great'' or ``terrible'' for \texttt{[MASK]}.
The objective of NSP further encourages BERT to learn contextual representation beyond word-level.
In the context of reviews, NSP formulates a task of ``artificial review prediction'', where a negative example is an original review but a positive example is a synthesized fake review by combining two different reviews.
This task exploits the rich relationships between two sides in the input, such as whether two sides of texts have the same rating or not (when two reviews with different ratings are combined as a positive example), or whether two sides are targeting the same product or not (when two reviews from different products are merged as a positive example).
In summary, these two objectives encourage to learn a myriad of fine-grained features for potential end tasks. 

We let the loss function of MLM be $\mathcal{L}_{\text{MLM}}$ and the loss function of next text piece prediction be $\mathcal{L}_{\text{NSP}}$, the total loss of the domain knowledge post-training is $\mathcal{L}_{\text{DK}}=\mathcal{L}_{\text{MLM}} + \mathcal{L}_{\text{NSP}} $.

To post-train BERT on task-aware knowledge, we use SQuAD (1.1), which is a popular large-scale MRC dataset.
Although BERT gains great success on SQuAD, this success is based on the huge amount of training examples of SQuAD (100,000+).
This amount is large enough to ameliorate the flaws of BERT that has almost no questions on the left side and no textual span predictions based on both the question and the document on the right side.
However, a small amount of fine-tuning examples is not sufficient to turn BERT to be more task-aware, as shown in Sec. \ref{sec:exp}.
We let the loss on SQuAD be $\mathcal{L}_{\text{MRC}}$, which is in a similar setting as the loss $\mathcal{L}_{\text{RRC}}$ for RRC.
As a result, the joint loss of post-training is defined as $\mathcal{L}=\mathcal{L}_{\text{DK}} + \mathcal{L}_{\text{MRC}}$.

One major issue of post-training on such a loss is the prohibitive cost of GPU memory usage.
Instead of updating parameters over a batch, we divide a batch into multiple sub-batches and accumulate gradients on those sub-batches before parameter updates. This allows for a smaller sub-batch to be consumed in each iteration.

\begin{algorithm}
\label{alg:pre-tuning}
\LinesNumbered
\DontPrintSemicolon
\caption{Post-training Algorithm}
\SetKwInOut{Input}{Input} 
\Input{$\mathcal{D}_\text{DK}$: one batch of DK data; \\$\mathcal{D}_\text{MRC}$ one batch of MRC data; \\$u$: number of sub-batches.}
\BlankLine
$\nabla_\Theta \mathcal{L} \gets 0 $ \;
$\{\mathcal{D}_{\text{DK}, 1}, \dots, \mathcal{D}_{\text{DK}, u} \} \gets \text{Split}(\mathcal{D}_\text{DK}, u) $ \;
$\{\mathcal{D}_{\text{MRC}, 1}, \dots, \mathcal{D}_{\text{MRC}, u} \} \gets \text{Split}(\mathcal{D}_\text{MRC}, u) $ \;
\For{$i \in \{1, \dots, u\}$ }{
    $\mathcal{L}_\text{partial}\gets \frac{\mathcal{L}_{\text{DK}}(\mathcal{D}_{\text{DK}, i}) + \mathcal{L}_{\text{MRC}}(\mathcal{D}_{\text{MRC}, i} )}{u} $ \;
    $\nabla_\Theta \mathcal{L} \gets \nabla_\Theta \mathcal{L} + \text{BackProp}(\mathcal{L}_\text{partial}) $\;
}
$\Theta \gets \text{ParameterUpdates}(\nabla_\Theta \mathcal{L}) $ \;
\end{algorithm}

Algorithm 1 describes one training step and takes one batch of data on domain knowledge (DK) $\mathcal{D}_\text{DK}$ and one batch of MRC training data $\mathcal{D}_\text{MRC}$ to update the parameters $\Theta$ of BERT.
In line 1, it first initializes the gradients $\nabla_\Theta$ of all parameters as 0 to prepare gradient computation. Then in lines 2 and 3, each batch of training data is split into $u$ sub-batches. Lines 4-7 spread the calculation of gradients to $u$ iterations, where the data from each iteration of sub-batches are supposed to be able to fit into GPU memory.
In line 5, it computes the partial joint loss $\mathcal{L}_\text{partial}$ of two sub-batches $\mathcal{D}_{\text{DK}, i}$ and $\mathcal{D}_{\text{MRC}, i}$ from the $i$-th iteration through forward pass.
Note that the summation of two sub-batches' losses is divided by $u$, which compensate the scale change introduced by gradient accumulation in line 6.
Line 6 accumulates the gradients produced by backpropagation from the partial joint loss. To this end, accumulating the gradients $u$ times is equivalent to computing the gradients on the whole batch once. But the sub-batches and their intermediate hidden representations during the $i$-th forward pass can be discarded to save memory space.
Only the gradients $\nabla_\Theta$ are kept throughout all iterations and used to update parameters (based on the chosen optimizer) in line 8.
We detail the hyper-parameter settings of this algorithm in Sec. \ref{sec:hyp}.

\section{Experiments}
\label{sec:exp}

We aim to answer the following research questions (RQs) in the experiment:

RQ1:~what is the performance gain of post-training for each review-based task, with respect to the state-of-the-art performance?

RQ2:~what is the performance of BERT's pre-trained weights on three review-based tasks without any domain and task adaptation?

RQ3:~upon ablation studies of separate domain knowledge post-training and task-awareness post-training, what is their respective contribution to the whole post-training performance gain?

\subsection{End Task Datasets}
As there are no existing datasets for RRC and
to be consistent with existing research on sentiment analysis, we adopt the \textit{laptop} and \textit{restaurant} reviews of SemEval 2016 Task 5 as the source to create datasets for RRC.
We do not use SemEval 2014 Task 4 or SemEval 2015 Task 12 because these datasets do not come with the review(document)-level XML tags to recover whole reviews from review sentences.~We keep the split of training and testing of the SemEval 2016 Task 5 datasets and annotate multiple QAs for each review following the way of constructing QAs for the SQuAD 1.1 datasets \cite{rajpurkar2016squad}.

To make sure our questions are close to real-world questions, 2 annotators are first exposed to 400 QAs from CQA (under the laptop category in Amazon.com or popular restaurants in Yelp.com) to get familiar with real questions.
Then they are asked to read reviews and independently label textual spans and ask corresponding questions when they feel the textual spans contain valuable information that customers may care about.
The textual spans are labeled to be as concise as possible but still human-readable.
Note that the annotations for sentiment analysis tasks are not exposed to annotators to avoid biased annotation on RRC.
Since it is unlikely that the two annotators can label the same QAs (the same questions with the same answer spans), they further mutually check each other's annotations and disagreements are discussed until agreements are reached.~Annotators are encouraged to label as many questions as possible from testing reviews to get more test examples. A training review is encouraged to have 2 questions (training examples) on average to have good coverage of reviews.

The annotated data is in the format of SQuAD 1.1 \cite{rajpurkar2016squad} to ensure compatibility with existing implementations of MRC models. The statistics of the RRC dataset (ReviewRC) are shown in Table \ref{tbl:rrc}. 
Since SemEval datasets do not come with a validation set, we further split 20\% of reviews from the training set for validation.

Statistics of datasets for AE and ASC are given in Table \ref{tbl:absa}.
For AE, we choose SemEval 2014 Task 4 for laptop and SemEval-2016 Task 5 for restaurant to be consistent with \cite{xu_acl2018} and other previous works.
For ASC, we use SemEval 2014 Task 4 for both laptop and restaurant as existing research frequently uses this version. We use 150 examples from the training set of all these datasets for validation.

\subsection{Post-training datasets}
For domain knowledge post-training, we use Amazon laptop reviews \cite{HeMcA16a} and Yelp Dataset Challenge reviews\footnote{\url{https://www.yelp.com/dataset/challenge}}.
For laptop, we filtered out reviewed products that have appeared in the validation/test reviews to avoid training bias for test data (Yelp reviews do not have this issue as the source reviews of SemEval are not from Yelp). 
Since the number of reviews is small, we choose a duplicate factor of 5 (each review generates about 5 training examples) during BERT data pre-processing.
This gives us 
1,151,863 post-training examples for laptop domain knowledge.

For the restaurant domain, we use Yelp reviews from restaurant categories that the SemEval reviews also belong to \cite{xu_acl2018}.
We choose 700K reviews to ensure it is large enough to generate training examples (with a duplicate factor of 1) to cover all post-training steps that we can afford (discussed in Section \ref{sec:hyp})\footnote{We expect that using more reviews can have even better results but we limit the amount of reviews based on our computational power.}.
This gives us 2,677,025 post-training examples for restaurant domain knowledge learning.

For MRC task-awareness  post-training, we leverage SQuAD 1.1 \cite{rajpurkar2016squad} that come with 87,599 training examples from 442 Wikipedia articles.

\subsection{Hyper-parameters}
\label{sec:hyp}
We adopt $\textbf{BERT}_\textbf{BASE}$ (uncased) as the basis for all experiments\footnote{We expect $\textbf{BERT}_\textbf{LARGE}$ to have better performance but leave that to future work due to limited computational power.}.~Since post-training may take a large footprint on GPU memory (as BERT pre-training), we leverage FP16 computation\footnote{\url{https://docs.nvidia.com/deeplearning/sdk/mixed-precision-training/index.html}} to reduce the size of both the model and hidden representations of data.~We set a static loss scale of 2 in FP16, which can avoid any over/under-flow of floating point computation.
The maximum length of post-training is set to 320 with a batch size of 16 for each type of knowledge.~The number of sub-batch $u$ is set to 2, which is good enough to store each sub-batch iteration into a GPU memory of 11G. We use Adam optimizer and set the learning rate to be 3e-5.
We train 70,000 steps for the laptop domain and 140,000 steps for the restaurant domain, which roughly have one pass over the pre-processed data on the respective domain.

\begin{table}
    \centering
    \scalebox{0.7}{
        \begin{tabular}{c||c|c}
        \hline
        {\bf Dataset} &{\bf Num. of Questions } &{\bf Num. of Reviews }  \\
        \hline
        Laptop Training & 1015 & 443 \\
        Laptop Testing & 351 & 79 \\
        \hline
        Restaurant Training & 799 & 347 \\
        Restaurant Testing & 431 & 90 \\
        \hline
        \end{tabular}
    }
	\caption{Statistics of the ReviewRC Dataset. Reviews with no questions are ignored.} 
\label{tbl:rrc}
\end{table}

\begin{table}
\centering
\scalebox{0.8}{
    \begin{tabular}{l|c|c}
    \hline
    {\bf } &{\bf AE} & {\bf ASC}\\
    \hline
    \bf{Laptop} & SemEval14 Task4 & SemEval14 Task4 \\
    \hline
    Training & 3045 S./2358 A. & 987 P./866 N./460 Ne. \\
    Testing & 800 S./654 A. & 341 P./128 N./169 Ne. \\
    \hline
    \bf{Restaurant} & SemEval16 Task5 & SemEval14 Task4 \\
    \hline
    Training & 2000 S./1743 A. & 2164 P./805 N./633 Ne. \\
    Testing & 676 S./622 A. & 728 P./196 N./196 Ne. \\
    \hline
    \end{tabular}
}
\caption{Summary of datasets on aspect extraction and aspect sentiment classification. S: number of sentences; A: number of aspects; P., N., and Ne.: number of positive, negative and neutral polarities.}
\label{tbl:absa}
\vspace{-5mm}
\end{table}

\subsection{Compared Methods}
As BERT outperforms existing open source MRC baselines by a large margin, we do not intend to exhaust existing implementations but focus on variants of BERT introduced in this paper.

\textbf{DrQA} is a baseline from the document reader\footnote{https://github.com/facebookresearch/DrQA} of DrQA \cite{chen2017reading}.~We adopt this baseline because of its simple implementation for reproducibility.~We run the document reader with random initialization and train it directly on ReviewRC.
We use all default hyper-parameter settings for this baseline except the number of epochs, which is set as 60 for better convergence.

\textbf{DrQA+MRC} is derived from the above baseline with official pre-trained weights on SQuAD.
We fine-tune document reader with ReviewRC. We expand the vocabulary of the embedding layer from the pre-trained model on ReviewRC since reviews may have words that are rare in Wikipedia and keep other hyper-parameters as their defaults.

For AE and ASC, we summarize the scores of the state-of-the-arts on SemEval (based the best of our knowledge) for brevity.\\
\textbf{DE-CNN} \cite{xu_acl2018} reaches the state-of-the-arts for AE by leveraging domain embeddings.\\
\textbf{MGAN} \cite{li2018exploiting} reaches the state-of-the-art ASC on SemEval 2014 task 4.

Lastly, to answer RQ1, RQ2, and RQ3, we have the following BERT variants.\\
\textbf{BERT} leverages the vanilla BERT pre-trained weights and fine-tunes on all 3 end tasks. We use this baseline to answer RQ2 and show that BERT's pre-trained weights alone have limited performance gains on review-based tasks.\\
\textbf{BERT-DK} post-trains BERT's weights only on domain knowledge (reviews) and fine-tunes on the 3 end tasks. We use BERT-DK and the following BERT-MRC to answer RQ3.\\
\textbf{BERT-MRC} post-trains BERT's weights on SQuAD 1.1 and then fine-tunes on the 3 end tasks.\\
\textbf{BERT-PT} (proposed method) post-trains BERT's weights using the joint post-training algorithm in Section \ref{sec:pt} and then fine-tunes on the 3 end tasks.

\subsection{Evaluation Metrics and Model Selection}
To be consistent with existing research on MRC,
we use the same evaluation script from SQuAD 1.1 \cite{rajpurkar2016squad} for RRC, which reports Exact Match (EM) and F1 scores.
EM requires the answers to have exact string match with human annotated answer spans.
F1 score is the averaged F1 scores of individual answers, which is typically higher than EM and is the major metric.
Each individual F1 score is the harmonic mean of individual precision and recall computed based on the number of overlapped words between the predicted answer and human annotated answers.

For AE, we use the standard evaluation scripts come with the SemEval datasets and report the F1 score.
For ASC, 
we compute both accuracy and Macro-F1 over 3 classes of polarities, where Macro-F1 is the major metric as the imbalanced classes introduce biases on accuracy.~To be consistent with existing research \cite{tang2016aspect}, examples belonging to the \textit{conflict} polarity are dropped due to a very small number of examples.

We set the maximum number of epochs to 4 for BERT variants, though most runs converge just within 2 epochs.
Results are reported as averages of \textbf{9} runs (9 different random seeds for random batch generation).\footnote{We notice that adopting 5 runs used by existing researches still has a high variance for a fair comparison.} 

\subsection{Result Analysis}

\begin{table}
    \centering
    \scalebox{0.7}{
        \begin{tabular}{l||c c|c c}
        \hline
        {\bf Domain} & {\bf Laptop} & & {\bf Rest.} & \\
        \hline
        {\bf Methods} & {\bf EM } &{\bf F1 } & {\bf EM } & {\bf F1 } \\
        \hline
        DrQA\cite{chen2017reading} & 38.26 & 50.99 & 49.52 & 63.73 \\
        DrQA+MRC\cite{chen2017reading} & 40.43 & 58.16 & 52.39 & 67.77 \\
        \hline
        BERT & 39.54 & 54.72 & 44.39 & 58.76 \\
        BERT-DK & 42.67 & 57.56 & 48.93 & 62.81 \\
        BERT-MRC &  47.01 & 63.87 & 54.78 & 68.84 \\
        BERT-PT & 48.05 & \textbf{64.51} & 59.22 & \textbf{73.08} \\
        \hline
        \end{tabular}
    }
	\caption{RRC in EM (Exact Match) and F1.}
\label{tbl:result_rc}
\vspace{-3mm}
\end{table}

\begin{table}
    \centering
    \scalebox{0.9}{
        \begin{tabular}{l||c|c}
        \hline
        {\bf Domain} & {\bf Laptop} & {\bf Rest.} \\
        \hline
        {\bf Methods} & {\bf F1 } & {\bf F1 } \\
        \hline
        \begin{tabular}{@{}l@{}}DE-CNN\cite{xu_acl2018}\end{tabular} & 81.59 & 74.37 \\
        \hline
        BERT  & 79.28 & 74.1 \\
        BERT-DK & 83.55 & 77.02 \\
        BERT-MRC & 81.06 & 74.21 \\
        BERT-PT & \textbf{84.26} & \textbf{77.97} \\
        \hline
        \end{tabular}
    }
	\caption{AE in F1.}
\label{tbl:result_ae}
\vspace{-5mm}
\end{table}

\begin{table}[t]
    \centering
    \scalebox{0.78}{
        \begin{tabular}{l||c c|c c}
        \hline
        {\bf Domain} & {\bf Laptop} & & {\bf Rest.} & \\
        \hline
        {\bf Methods} & \bf{Acc.} & \bf{MF1} & \bf{Acc.} & \bf{MF1} \\
        \hline
        \begin{tabular}{@{}l@{}}
        MGAN \cite{li2018exploiting}\end{tabular} & 76.21 & 71.42 & 81.49 & 71.48 \\
        \hline
        BERT & 75.29 & 71.91 & 81.54 & 71.94 \\
        BERT-DK & 77.01 & 73.72 & 83.96 & 75.45 \\
        BERT-MRC & 77.19 & 74.1 & 83.17 & 74.97 \\
        BERT-PT & 78.07 & \textbf{75.08} & 84.95 & \textbf{76.96} \\
        \hline
        \end{tabular}
    }
	\caption{ASC in Accuracy and Macro-F1(MF1).}
\label{tbl:result_asc}
\vspace{-5mm}
\end{table}

The results of RRC, AE and ASC are shown in Tables \ref{tbl:result_rc}, \ref{tbl:result_ae} and \ref{tbl:result_asc}, respectively. To answer RQ1, we observed that the proposed joint post-training (BERT-PT) has the best performance over all tasks in all domains, which show the benefits of having two types of knowledge.

To answer RQ2, to our surprise we found that the vanilla pre-trained weights of BERT do not work well for review-based tasks, although it achieves state-of-the-art results on many other NLP tasks \cite{devlin2018bert}.
This justifies the need to adapt BERT to review-based tasks.

To answer RQ3, we noticed that the roles of domain knowledge and task knowledge vary for different tasks and domains.
For RRC, we found that the performance gain of BERT-PT mostly comes from task-awareness (MRC) post-training (as indicated by BERT-MRC).
The domain knowledge helps more for restaurant than for laptop.
We suspect the reason is that certain types of knowledge (such as specifications) of laptop are already present in Wikipedia, whereas Wikipedia has little knowledge about restaurant.
We further investigated the examples improved by BERT-MRC and found that the boundaries of spans (especially short spans) were greatly improved. 

For AE,
we found that great performance boost comes mostly from domain knowledge post-training, which indicates that contextualized representations of domain knowledge are very important for AE. BERT-MRC has almost no improvement on restaurant, which indicates Wikipedia may have no knowledge about aspects of restaurant.
We suspect that the improvements on laptop come from the fact that many answer spans in SQuAD are noun terms, which bear a closer relationship with laptop aspects.

For ASC, we observed that large-scale annotated MRC data is very useful.
We suspect the reason is that ASC can be interpreted as a special MRC problem, where all questions are about the polarity of a given aspect.
MRC training data may help BERT to understand the input format of ASC given their closer input formulation.
Again, domain knowledge post-training also helps ASC.

We further investigated the errors from BERT-PT over the 3 tasks.
The errors on RRC mainly come from boundaries of spans that are not concise enough and incorrect location of spans that may have certain nearby words related to the question. 
We believe precisely understanding user's experience is challenging from only domain post-training given limited help from the RRC data and no help from the Wikipedia data.
For AE, errors mostly come from annotation inconsistency and boundaries of aspects (e.g., apple OS is predicted as OS). Restaurant suffers from rare aspects like the names of dishes.
ASC tends to have more errors as the decision boundary between the negative and neutral examples is unclear (e.g., even annotators may not sure whether the reviewer shows no opinion or slight negative opinion when mentioning an aspect).
Also, BERT-PT has the problem of dealing with one sentence with two opposite opinions (``The screen is good but not for windows.''). We believe that such training examples are rare.

\section{Conclusions}
We proposed a new task called \textit{review reading comprehension} (RRC) and investigated the possibility of turning reviews as a valuable resource for answering user questions.
We adopted BERT as our base model and proposed a joint post-training approach to enhancing both the domain and task knowledge.~We further explored the use of this approach in two other review-based tasks: aspect extraction and aspect sentiment classification. 
Experimental results show that the post-training approach before fine-tuning is effective.

\section*{Acknowledgments}
Bing Liu's work was partially supported by the National Science
Foundation (NSF IIS 1838770) and by a research gift from Huawei.

\bibliography{naaclhlt2019}
\bibliographystyle{acl_natbib}

\end{document}